# A Sound and Complete Algorithm for Learning Causal Models from Relational Data

Marc Maier    Katerina Marazopoulou    David Arbour    David Jensen

School of Computer Science
University of Massachusetts Amherst
Amherst, MA 01003
{maier, kmarazo, darbour, jensen}@cs.umass.edu

## Abstract

The PC algorithm learns maximally oriented causal Bayesian networks. However, there is no equivalent complete algorithm for learning the structure of relational models, a more expressive generalization of Bayesian networks. Recent developments in the theory and representation of relational models support lifted reasoning about conditional independence. This enables a powerful constraint for orienting bivariate dependencies and forms the basis of a new algorithm for learning structure. We present the *relational causal discovery* (RCD) algorithm that learns causal relational models. We prove that RCD is sound and complete, and we present empirical results that demonstrate effectiveness.

## 1 INTRODUCTION

Research in causal discovery has led to the identification of fundamental principles and methods for causal inference, including a *complete* algorithm—the PC algorithm—that identifies all possible orientations of causal dependencies from observed conditional independencies (Pearl, 2000; Spirtes et al., 2000; Meek, 1995). Completeness guarantees that no other method can infer more causal dependencies from observational data. However, much of this work, including the completeness result, applies only to Bayesian networks.

Over the past 15 years, researchers have developed more expressive classes of models, including probabilistic relational models (Getoor et al., 2007), that remove the assumption of independent and identically distributed instances required by Bayesian networks. These *relational* models represent systems involving multiple types of interacting entities with probabilistic dependencies among them. Most algorithms for learning the structure of relational models focus on statistical association. The single algorithm that does address causality—Relational PC (Maier et al., 2010)—is not complete and is prone to orientation errors, as we show in this paper. Consequently, there is no relational analog to the completeness result for Bayesian networks.

Recent advances in the theory and representation of relational models provide a foundation for reasoning about causal dependencies (Maier et al., 2013). That work develops a novel, lifted representation—the *abstract ground graph*—that abstracts over all instantiations of a relational model, and it uses this abstraction to develop the theory of relational $d$-separation. This theory connects the causal structure of a relational model and probability distributions, similar to how $d$-separation connects the structure of Bayesian networks and probability distributions.

We present the implications of abstract ground graphs and relational $d$-separation for learning causal models from relational data. We introduce a powerful constraint that can orient bivariate dependencies (yielding models with up to 72% additional oriented dependencies) without assumptions on the underlying distribution. We prove that this new rule, called *relational bivariate orientation*, combined with relational extensions to the rules utilized by the PC algorithm, yields a sound and complete approach to identifying the causal structure of relational models. We develop a new algorithm, called *relational causal discovery* (RCD), that leverages these constraints, and we prove that RCD is sound and complete under the causal sufficiency assumption. We show RCD's effectiveness with a practical implementation and compare it to several alternative algorithms. Finally, we demonstrate RCD on a real-world dataset drawn from the movie industry.

## 2 EXAMPLE

Consider a data set containing actors with a measurement of their popularity (e.g., price on the Hollywood Stock Exchange) and the movies they star in with a measurement of success (e.g., box office revenue). A simple analysis might detect a statistical association between popularity and success, but the models in which popularity causes success and success causes popularity may be statistically indistinguishable.

However, with knowledge of the *relational structure*, a considerable amount of information remains to be leveraged. From the perspective of actors, we can ask whether one actor's popularity is conditionally independent of the popularity of other actors appearing in the same movie, given that movie's success. Similarly, from the perspective of movies, we can ask whether the success of a movie is conditionally independent of the success of other movies with a common actor, given that actor's popularity. With conditional independence, we now can determine the orientation for a single *relational dependency*.

These additional tests of conditional independence manifest when inspecting relational data with *abstract ground graphs*—a lifted representation developed by Maier et al. (2013) (see Section 3.2 for more details). If actor popularity indeed causes movie success, then the popularity of actors appearing in the same movie would be marginally independent. This produces a collider from the actor perspective and a common cause from the movie perspective, as shown in Figure 1. With this representation, it is straightforward to identify the orientation of such a bivariate dependency.

This example illustrates two central ideas of this paper. First, abstract ground graphs enable a new constraint on the space of causal models—relational bivariate orientation. The rules used by the PC algorithm can also be adapted to orient the edges of abstract ground graphs (Section 4). Second, this constraint-based approach—testing for conditional independencies and reasoning about them to orient causal dependencies—is the primary strategy of the relational causal discovery algorithm (Section 5).

## 3 BACKGROUND

The details of RCD and its correctness rely on fundamental concepts of relational data, models, and *d*-separation as provided by Maier et al. (2013). This section provides a review of this theory in the context of the movie domain example. Note that the relational representation is a strictly more general framework for causal discovery, reducing to Bayesian networks in the presence of a single entity with no relationships.

### 3.1 RELATIONAL DATA AND MODELS

A *relational schema*, $\mathcal{S} = (\mathcal{E}, \mathcal{R}, \mathcal{A})$, describes the entity, relationship, and attribute classes in a domain, as well as cardinality constraints for the number of entity instances involved in a relationship. A schema is typically depicted with an entity-relationship diagram, such as the one underlying the model shown in Figure 2(a). This example has two entity classes—ACTOR with attribute *Popularity* and MOVIE with attribute *Success*—and one relationship class—STARS-IN with no attributes. The cardinality constraints (expressed as crow's feet in the diagram) indicate that many actors may appear in a movie and a single actor may appear in many movies. A schema is a template for a *relational skeleton $\sigma$*—a data set of entity and relationship instances. The example in Figure 2(b) contains four ACTOR instances, five MOVIE instances, and the relationships among them.

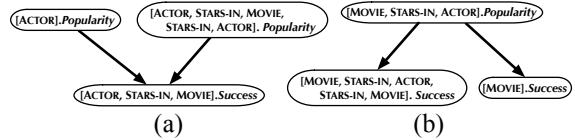

Figure 1: Abstract ground graphs from (a) the ACTOR perspective and (b) the MOVIE perspective.

Given a relational schema, one can specify *relational paths*, which are critical for specifying the variables and dependencies of a relational model. A relational path is an alternating sequence of entity and relationship classes that follow connected paths in the schema (subject to cardinality constraints). In Figure 2(a), possible relational paths include [ACTOR] (a singleton path specifying an actor), [MOVIE, STARS-IN, ACTOR] (specifying the actors in a movie), or even [ACTOR, STARS-IN, MOVIE, STARS-IN, ACTOR] (describing co-stars). The cardinality of a relational path is MANY if the cardinalities along the path indicate that it could reach more than one instance; otherwise, the cardinality is ONE. For example, $card$([MOVIE, STARS-IN, ACTOR]) = MANY since a movie can reach many actors, whereas $card$([ACTOR]) = ONE since this path can only reach the base actor instance.

*Relational variables* consist of a relational path and an attribute, and they describe attributes of classes reached via a relational path (e.g., the popularity of actors starring in a movie). *Relational dependencies* consist of a pair of relational variables with a common first item, called the *perspective*. The dependency in Figure 2(a) states that the popularity of actors influences the success of movies they star in. A *canonical* dependency has a single item class in the relational path of the effect variable. A *relational model*, $\mathcal{M} = (\mathcal{S}, \mathcal{D})$, is a collection of relational dependencies $\mathcal{D}$, in canonical form, defined over schema $\mathcal{S}$. Relational models are parameterized by a set of conditional probability dis-tributions, one for each attribute class $\mathcal{A}(I)$ for each $I \in \mathcal{E} \cup \mathcal{R}$, that factorizes a joint probability distribution for a given skeleton. This class of models can be expressed as DAPER models (Heckerman et al., 2007), and they are more general than plate models (Buntine, 1994; Gilks et al., 1994) and than PRMs with dependencies among only attributes (Getoor et al., 2007).

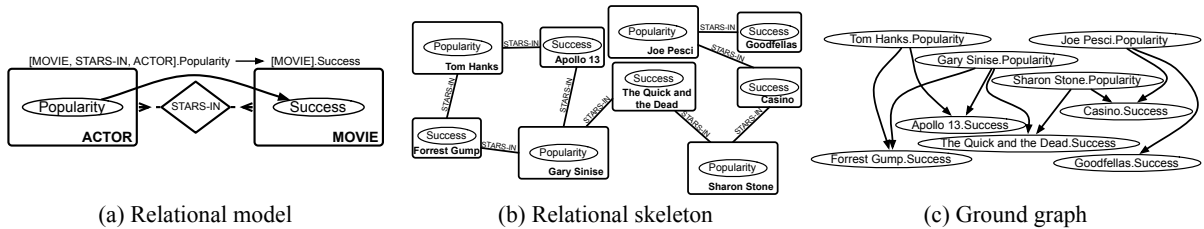

(a) Relational model    (b) Relational skeleton    (c) Ground graph

Figure 2: An example relational model involving actors and movies with a single relational dependency stating that actor popularity causes movie success. The variables in the ground graph are drawn from the instances in the skeleton, and the dependencies in the ground graph are drawn from the dependency in the model.

A relational model $\mathcal{M}$ paired with a relational skeleton $\sigma$ produces a model instantiation $GG_{\mathcal{M}\sigma}$, called the *ground graph*. A ground graph is a directed graph with a node for each attribute of every entity and relationship instance in $\sigma$, and an edge between instances of relational variables for all dependencies in $\mathcal{M}$. A single relational model is a template for all possible ground graphs, one for every possible skeleton. Figure 2(c) shows an example ground graph. A ground graph has the same semantics as a Bayesian network with joint probability $P(GG_{\mathcal{M}\sigma}) = \prod_{I \in \mathcal{E} \cup \mathcal{R}} \prod_{X \in \mathcal{A}(I)} \prod_{i \in \sigma(I)} P(i.X \mid parents(i.X))$.

### 3.2 ABSTRACT GROUND GRAPHS

The RCD algorithm reasons about conditional independence using *abstract ground graphs*, introduced by Maier et al. (2013). Unlike the reasoning it supports in Bayesian networks, $d$-separation does not accurately infer conditional independence when applied directly to relational models. Abstract ground graphs enable sound and complete derivation of conditional independence facts using $d$-separation.

An abstract ground graph $AGG_{\mathcal{M}Bh}$ for relational model $\mathcal{M}$, perspective $B \in \mathcal{E} \cup \mathcal{R}$, and hop threshold $h \in \mathbb{N}^0$ is a directed graph that captures the dependencies among relational variables holding for any possible ground graph. $AGG_{\mathcal{M}Bh}$ has a node for each relational variable from perspective $B$ with path length limited by $h$. $AGG_{\mathcal{M}Bh}$ contains edges between relational variables if the instantiations of those relational variables contain a dependent pair in some ground graph. Note that a *single* dependency in $\mathcal{M}$ may support *many* edges in $AGG_{\mathcal{M}Bh}$. Additionally, a *single* model $\mathcal{M}$ may produce *many* abstract ground graphs, one for each perspective.

Figure 1 shows abstract ground graphs for the model in Figure 2(a) from the ACTOR and MOVIE perspectives with $h = 4$. There is a single relational dependency in the example model, yet it supports two edges in each abstract ground graph. Also, one perspective exhibits a collider while the other contains a common cause. The abstract ground graph is the underlying representation used by RCD, and the conditional independence facts derived from it form the crux of the relational bivariate orientation rule.

## 4 EDGE ORIENTATION

Edge orientation rules, such as those used by the PC algorithm, use patterns of dependence and conditional independence to determine the direction of causality (Spirtes et al., 2000). In this section, we present the relational bivariate orientation rule and describe how the PC orientation rules can orient the edges of abstract ground graphs. We also prove that these orientation rules are individually sound and collectively complete for causally sufficient relational data.

### 4.1 BIVARIATE EDGE ORIENTATION

The example from Section 2 briefly describes the application of relational bivariate orientation (RBO). The abstract ground graph representation presents an opportunity to orient dependencies that cross relationships with a MANY cardinality. RBO requires no assumptions about functional form or conditional densities, unlike the recent work by Shimizu et al. (2006), Hoyer et al. (2008), and Peters et al. (2011) to orient bivariate dependencies. The only required assumption is the standard model acyclicity assumption, which restricts the space of dependencies to those without direct or indirect feedback cycles.

In the remainder of the paper, let $I_W$ denote the item class on which attribute $W$ is defined, and let $X - Y$ denote an undirected edge.

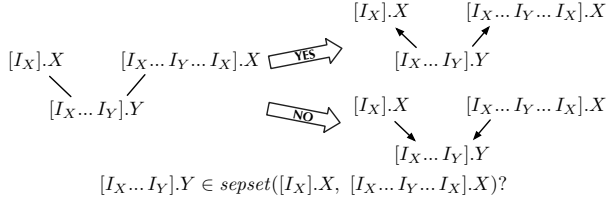

$[I_X \ldots I_Y].Y \in sepset([I_X].X, \ [I_X \ldots I_Y \ldots I_X].X)?$

Figure 3: The relational bivariate orientation rule is conditional on whether $[I_X \ldots I_Y].Y$ is in the separating set of $[I_X].X$ and $[I_X \ldots I_Y \ldots I_X].X$.

**Definition 1 (Relational Bivariate Orientation)**
Let $\mathcal{M}$ be a relational model and $G$ a partially directed abstract ground graph for $\mathcal{M}$, perspective $I_X$, and hop threshold $h$. If $[I_X].X - [I_X \ldots I_Y].Y$ is in $G$, $card([I_Y \ldots I_X]) =$ MANY, and $[I_X].X \perp\!\!\!\perp [I_X \ldots I_Y \ldots I_X].X \mid \mathbf{Z}$, then (1) if $[I_X \ldots I_Y].Y \in \mathbf{Z}$, orient as $[I_X].X \leftarrow [I_X \ldots I_Y].Y$; (2) if $[I_X \ldots I_Y].Y \notin \mathbf{Z}$, orient as $[I_X].X \rightarrow [I_X \ldots I_Y].Y$.

RBO is illustrated in Figure 3. Given Definition 1, if $[I_X \ldots I_Y].Y$ is a collider for perspective $I_X$, then $[I_Y \ldots I_X].X$ is a common cause for perspective $I_Y$, assuming $card([I_Y \ldots I_X]) =$ MANY $= card([I_X \ldots I_Y])$. If $card([I_X \ldots I_Y]) =$ ONE and $card([I_Y \ldots I_X]) =$ MANY, then RBO applies only to the abstract ground graph with perspective $I_X$. For the example in Figure 1(a), [ACTOR, STARS-IN, MOVIE].$Success$ is a collider for the ACTOR perspective.

RBO is akin to detecting relational autocorrelation (Jensen and Neville, 2002) and checking whether a distinct variable is a member of the set that eliminates the autocorrelation. It is also different than the collider detection rule (see Section 4.2) because it can explicitly orient dependencies as a common cause when the unshielded triple does not present itself as a collider. In Section 6.1, we quantify the extent to which RBO provides additional information beyond the standard PC edge orientation rules.

### 4.2 ORIENTING THE EDGES OF ABSTRACT GROUND GRAPHS

We adapt the rules for orienting edges in a Bayesian network, as used by PC (Spirtes et al., 2000) and characterized theoretically by Meek (1995), to orient relational dependencies at the level of abstract ground graphs. Figure 4 displays the four rules[1]—Collider Detection (CD), Known Non-Colliders (KNC), Cycle Avoidance (CA), and Meek Rule 3 (MR3)—as they would appear in an abstract ground graph.

---
[1] An additional rule is described by Meek (1995), but it only activates given prior knowledge.

A relational model has a corresponding set of abstract ground graphs, one for each perspective, but all are derived from the same relational dependencies. Recall from Section 3.2 that a single dependency supports many edges within and across the set of abstract ground graphs. Consequently, when a rule is activated for a *specific* abstract ground graph, the orientation of the underlying relational dependency must be propagated within and across *all* abstract ground graphs.

### 4.3 PROOF OF SOUNDNESS

An orientation rule is *sound* if any orientation not indicated by the rule introduces either (1) an unshielded collider in some abstract ground graph, (2) a directed cycle in some abstract ground graph, or (3) a cycle in the relational model (adapted from the definition of soundness given by Meek (1995)).

**Theorem 1** *Let $G$ be a partially oriented abstract ground graph from perspective $B$ with correct adjacencies and correctly oriented unshielded colliders by either CD or RBO. Then, KNC, CA, MR3, and the purely common cause case of RBO, as well as the embedded orientation propagation, are sound.*

**Proof.** The proof for KNC, CA, and MR3 is nearly identical to the proof given by Meek (1995).

Orientation propagation: Let $[B \ldots I_X].X \rightarrow [B \ldots I_Y].Y$ be an oriented edge in $G$. By the definition of abstract ground graphs, this edge stems from a relational dependency $[I_Y \ldots I_X].X \rightarrow [I_Y].Y$. Let $[B \ldots I_X]'.X - [B \ldots I_Y].Y$ be an unoriented edge in $G$ where $[B \ldots I_X]'$ is different than $[B \ldots I_X]$, but the edge is supported by the same underlying relational dependency. Assume for contradiction that the edge is oriented as $[B \ldots I_X]'.X \leftarrow [B \ldots I_Y].Y$. Then, there must exist a dependency $[I_X \ldots I_Y].Y \rightarrow [I_X].X$ in the model, which yields a cycle. The argument is the same for abstract ground graphs from different perspectives.

RBO common cause case: Given Definition 1, no alternate perspective would have oriented the triple as a collider, and $B = I_X$. Let $[I_X].X - [I_X \ldots I_Y].Y - [I_X \ldots I_Y \ldots I_X].X$ be an unoriented triple in $G$. Assume for contradiction that the triple is oriented as $[I_X].X \rightarrow [I_X \ldots I_Y].Y \leftarrow [I_X \ldots I_Y \ldots I_X].X$. This creates a new unshielded collider. Assume for contradiction that the triple is oriented as $[I_X].X \rightarrow [I_X \ldots I_Y].Y \rightarrow [I_X \ldots I_Y \ldots I_X].X$ or equivalently, the reverse direction. This implies a cycle in the model. ∎

### 4.4 PROOF OF COMPLETENESS

A set of orientation rules is complete if it produces a maximally oriented graph. Any orientation of an un-oriented edge must be consistent with a member of the Markov equivalence class. Lemma 1 describes a useful property that enables the proof of completeness to reason directly about the remaining unoriented edges.

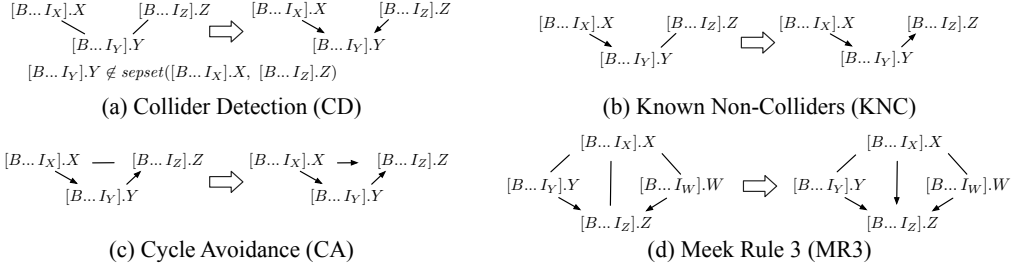

Figure 4: Schematics of the PC orientation rules on an abstract ground graph from perspective $B$.

**Lemma 1** *Let $G$ be a partially oriented abstract ground graph, with correct adjacencies and oriented unshielded colliders. Let $G_o$ be the result of exhaustively applying KNC, CA, MR3, and the purely common cause case of RBO all with orientation propagation. In $G_o$, if $P.X \rightarrow P'.Y - P''.Z$, then $P.X \rightarrow P''.Z$.*

**Proof.** Much of this proof follows from Meek (1995).

The following properties hold: (1) $X \neq Z$; otherwise, RBO would have oriented $P'.Y \leftarrow P''.Z$. (2) $P.X$ must be adjacent to $P''Z$; otherwise, KNC would have oriented $P'.Y \rightarrow P''.Z$. (3) $P.X \leftarrow P''.Z$ *does not* hold; otherwise, CA would have oriented $P'.Y \leftarrow P''.Z$. Therefore, we have a structure of the form $P.X \rightarrow P'.Y - P''.Z$ and $P.X - P''.Z$.

We show that $P.X \rightarrow P''.Z$ through exhaustive enumeration of the cases under which $P.X \rightarrow P'.Y$ was oriented. The cases for KNC, CD (and RBO collider cases), CA, and MR3 follow directly from Meek (1995).

(1) RBO oriented $P.X \rightarrow P'.Y$ from the $I_Y$ perspective as a common cause. Then, $P' = [I_Y]$, $P = [I_Y ... I_X]$, and $P'' = [I_Y ... I_Z]$. Also, $[I_Y ... I_X ... I_Y].Y$ must be in $G_o$ with $[I_Y ... I_X].X \rightarrow [I_Y ... I_X ... I_Y].Y$. By Definition 1, $card([I_Y ... I_X]) =$ ONE and $card([I_Y ... I_X]) =$ MANY.

The relational path $[I_Y ... I_Z]$ and its reverse have cardinality ONE; otherwise, RBO would have oriented $[I_Y].Y - [I_Y ... I_Z].Z$. We show that $[I_Y ... I_X].X - [I_Y ... I_Z].Z$ cannot remain unoriented.

Since this edge exists, by the construction of abstract ground graphs, (a) $[I_Y ... I_X]$ must be produced by combining $[I_Y ... I_Z]$ and $[I_Z ... I_X]$) (using the *extend* method (Maier et al., 2013)) and (b) $[I_Y ... I_Z]$ must be produced by combining $[I_Y ... I_X]$ and $[I_X ... I_Z]$). The paths $[I_X ... I_Z]$ and $[I_Z ... I_X]$ underlie the dependency between $X$ and $Z$. Facts (a) and (b) impose constraints on the schema and abstract ground graphs. There are four cases for (a) depending on the relationship between $[I_Y ... I_Z]$ and $[I_Z ... I_X]$, with equivalent cases for (b).

(i) $[I_Y ... I_Z]$ and $[I_Z ... I_X]$ overlap exactly at $I_Z$. Then, the path from $I_X$ to $I_Z$ must have cardinality MANY. This implies that from the $I_Z$ perspective, RBO would have oriented $X$ to $Z$.

(ii) $[I_Y ... I_M ... I_Z]$ and $[I_Z ... I_M ... I_X]$ overlap up to $I_M$. This is equivalent to case (i), except $I_M$ appears on the path from $I_X$ to $I_Z$.

(iii) $[I_Z ... I_X]$ is a subpath of the reverse of $[I_Y ... I_Z]$. Then, the path from $I_Z$ to $I_Y$ must have cardinality MANY, which is a contradiction.

(iv) The reverse of $[I_Y ... I_Z]$ is a subpath of $[I_Z ... I_X]$. This is equivalent to case (i), except $I_Y$ appears on the path from $I_X$ to $I_Z$.

(2) Orientation propagation oriented $P.X \rightarrow P'.Y$. Then, there exists an edge for some perspective that was oriented by one of the orientation rules. From that perspective, the local structure matches the given pattern, and from the previous cases, $X \rightarrow Z$ was oriented. By definition, $P.X \rightarrow P''.Z$. ∎

Meek (1995) also provides the following results, used for proving completeness. A *chordal graph* is an undirected graph where every undirected cycle of length four or more has an edge between two nonconsecutive vertices on the cycle. Let $G$ be an undirected graph, $\alpha$ a total order on the vertices of $G$, and $G_\alpha$ the induced directed graph ($A \rightarrow B$ is in $G_\alpha$ if and only if $A < B$ with respect to $\alpha$). A total order $\alpha$ is *consistent* with respect to $G$ if and only if $G_\alpha$ has no unshielded colliders. It can be shown that only chordal graphs have consistent orderings. Finally, if $G$ is an undirected chordal graph, then for all pairs of adjacent vertices $A$ and $B$ in $G$, there exist consistent total orderings $\alpha$ and $\gamma$ such that $A \rightarrow B$ in $G_\alpha$ and $A \leftarrow B$ in $G_\gamma$.

**Theorem 2** *Given a partially oriented abstract ground graph, with correct adjacencies and oriented*

unshielded colliders, exhaustively applying KNC, CA, MR3, and RBO all with orientation propagation results in a maximally oriented graph $G$.

**Proof.** Much of this proof follows from Meek (1995). Let $E_u$ and $E_o$ be the set of unoriented edges and oriented edges of $G$, respectively.

*Claim 1:* No orientation of edges in $E_u$ creates a cycle or unshielded collider in $G$ that includes edges from $E_o$.
*Proof.* Assume there exists an orientation of edges in $E_u$ that creates a cycle using edges from $E_o$. Without loss of generality, assume that the cycle is of length three. (1) If $A \rightarrow B \rightarrow C$ are in $E_o$ and $A-C$ in $E_u$, then CA would have oriented $A \rightarrow C$. (2) If $A \rightarrow B \leftarrow C$ or $A \leftarrow B \rightarrow C$ are in $E_o$ and $A-C$ is in $E_u$, then no orientation $A-C$ would create a cycle. (3) If $A \rightarrow B$ is in $E_o$ and $B-C-A$ in $E_u$, then by Lemma 1 we have $A \rightarrow C$ and no orientation of $B-C$ would create a cycle. A similar argument holds for unshielded colliders. □

*Claim 2:* Let $G_u$ be the subgraph of $G$ containing only unoriented edges. $G_u$ is the union of disjoint chordal graphs.
*Proof.* Assume that $G_u$ is not the union of disjoint chordal graphs. Then, there exists at least one disjoint component of $G_u$ that is not a chordal graph. Recall that no total ordering of $G_u$ is consistent. Let $A \rightarrow B \leftarrow C$ be an unshielded collider induced by some ordering on $G_u$. There are two cases: (1) $A$ and $C$ are adjacent in $G$. The edge must be oriented; otherwise, it would appear in $G_u$. Both orientations of $A-C$ imply an orientation of $A$ and $B$, or $C$ and $B$, by Lemma 1. (2) $A$ and $C$ are not adjacent in $G$. Then, $A-B-C$ is an unshielded triple in $G$. Either CD or RBO would have oriented the triple as a collider, or the triple is inconsistent with the total ordering on $G_u$. □

Since $G$ is chordal, it follows that no orientation of the unoriented edges in $G$ creates a new unshielded collider or cycle. ∎

## 5 The RCD Algorithm

The relational causal discovery (RCD) algorithm is a sound and complete constraint-based algorithm for learning causal models from relational data.[2] RCD employs a similar strategy to the PC algorithm, operating in two distinct phases (Spirtes et al., 2000). RCD is similar to the Relational PC (RPC) algorithm, which also learns causal relational models (Maier et al., 2010). The differences between RPC and RCD are threefold: (1) The underlying representation for RCD is a set of abstract ground graphs; (2) RCD employs a new causal constraint—the relational bivariate orientation rule; and (3) RCD is sound and complete. RPC also reasons about the uncertainty of relationship existence, but RCD assumes a prior relational skeleton.

**ALGORITHM 1:** RCD($schema, depth, hopThreshold, P$)
1   $PDs \leftarrow$ getPotentialDeps($schema, hopThreshold$)
2   $N \leftarrow$ initializeNeighbors($schema, hopThreshold$)
3   $S \leftarrow \{\}$
  // Phase I
4   **for** $d \leftarrow 0$ **to** $depth$ **do**
5     **for** $X \rightarrow Y \in PDs$ **do**
6       **foreach** $condSet \in$ powerset($N[Y] \setminus \{X\}$) **do**
7         **if** $|condSet| = d$ **then**
8           **if** $X \perp\!\!\!\perp Y \mid condSet$ in $P$ **then**
9             $PDs \leftarrow PDs \setminus \{X \rightarrow Y, Y \rightarrow X\}$
10            $S[X,Y] \leftarrow condSet$
11            **break**
  // Phase II
12   $AGGs \leftarrow$ buildAbstractGroundGraph($PDs$)
13   $AGGs, S \leftarrow$ ColliderDetection($AGGs, S$)
14   $AGGs, S \leftarrow$ BivariateOrientation($AGGs, S$)
15   **while** *changed* **do**
16     $AGGs \leftarrow$ KnownNonColliders($AGGs, S$)
17     $AGGs \leftarrow$ CycleAvoidance($AGGs, S$)
18     $AGGs \leftarrow$ MeekRule3($AGGs, S$)
19   **return** getCanonicalDependencies($AGGs$)

The remainder of this section describes the algorithmic details of RCD and proves its correctness.

Algorithm 1 provides pseudocode for RCD. Initially, RCD enumerates the set of potential dependencies, in canonical form, with relational paths limited by the hop threshold (line 1). Phase I continues similarly to PC, removing potential dependencies via conditional independence tests with conditioning sets of increasing size drawn from the power set of neighbors of the effect variable (lines 4–11). Every identified separating set is recorded, and the corresponding potential dependency and its reverse are removed (lines 9–10).

The second phase of RCD determines the orientation of dependencies consistent with the conditional independencies discovered in Phase I. First, Phase II constructs a set of undirected abstract ground graphs, one for each perspective, given the remaining dependencies. RCD then iteratively checks all edge orientation rules, as described in Section 4. Phase II of RCD is also different from PC and RPC because it searches for additional separating sets while finding colliders and common causes with CD and RBO. Frequently, unshielded triples $X-Y-Z$ may have no separating set recorded for $X$ and $Z$. For these pairs, RCD attempts to discover a new separating set, as in Phase I. These triples occur for one of three reasons: (1) Since $X$ and $Z$ are relational variables, the separating set may have been discovered from an alternative perspective; (2) The total number of hops in the relational paths for $X$, $Y$, and $Z$ may exceed the hop threshold—each dependency is subject to the hop threshold, but a pair of dependencies is limited by twice the hop threshold; or (3) The attributes of relational variables $X$ and $Z$ are the same, which is necessarily excluded as a potential dependency by the assumption of an acyclic model.

---
[2] Code available at kdl.cs.umass.edu/rcd.

Given the algorithm description and the soundness and completeness of the edge orientation rules, we prove that RCD is sound and complete. The proof assumes causal sufficiency and a prior relational skeleton (i.e., no causes of the relational structure).

**Theorem 3** *Given a schema and probability distribution P, RCD learns a correct maximally oriented model $\mathcal{M}$ assuming perfect conditional independence tests, sufficient hop threshold h, and sufficient depth.*

**Proof sketch.** Given sufficient $h$, the set of potential dependencies $PDs$ includes all true dependencies in $\mathcal{M}$, and the set of neighbors $N$ includes the true causes for every effect relational variable. Assuming perfect conditional independence tests, $PDs$ includes exactly the undirected true dependencies after Phase I, and $S[X, Y]$ records a correct separating set for the relational variable pair $\langle X, Y \rangle$. However, there may exist non-adjacent pairs of variables that have no recorded separating set (for the three reasons mentioned above). Given the remaining dependencies in $PDs$, we construct the correct set of edges in $AGGs$ using the methods from Maier et al. (2013). Next, all unshielded colliders are oriented by either CD or RBO, with correctness following from Spirtes et al. (2000) and relational $d$-separation (Maier et al., 2013). Whenever a pair $\langle X, Y \rangle$ is missing a separating set in $S$, it is either found as in Phase I or from a different perspective. RCD then produces a maximally oriented model by the soundness (Theorem 1) and completeness (Theorem 2) results of the remaining orientation rules. ∎

## 6 EXPERIMENTS

### 6.1 SYNTHETIC EXPERIMENTS

The proofs of soundness and completeness offer a qualitative measure of RCD's effectiveness—no other method can learn a more accurate causal model from observational data. To complement the theoretical results, we provide a *quantitative* measure of RCD's performance and compare against the performance of alternative constraint-based algorithms.

We evaluate RCD against two alternative algorithms. The first algorithm is RPC (Maier et al., 2010). This provides a comparison against current state-of-the-art relational learning. The second algorithm is the PC algorithm executed on relational data that has been propositionalized from a specific perspective—termed Propositionalized PC (PPC). Propositionalization reduces relational data to a single, propositional table (Kramer et al., 2001). We take the best and worst perspectives for each trial by computing the average F-score of its skeleton and compelled models.

We generated 1,000 random causal models over randomly generated schemas for each of the following combinations: entities (1–4); relationships (one less than the number of entities) with cardinalities selected uniformly at random; attributes per item drawn from $Pois(\lambda = 1) + 1$; and relational dependencies (1–15) limited by a hop threshold of 4 and at most 3 parents per variable. This procedure yielded a total of 60,000 synthetic models. Note that this generates simple Bayesian networks when there is a single entity class. We ran RCD, RPC, and PPC for each perspective, using a relational $d$-separation oracle with hop threshold 8 for the abstract ground graphs.

We compare the learned causal models with the true causal model. For each trial, we record the precision (the proportion of learned edges in the true model) and recall (the proportion of true edges in the learned model) for both the undirected skeleton after Phase I and the partially orientated model after Phase II. Figure 5 displays the average across 1,000 trials for each algorithm and measure. We omit error bars as the maximum standard error was less than 0.015.

All algorithms learn identical models for the single-entity case because they reduce to PC when analyzing propositional data. For truly relational data, algorithms that reason over relational representations are necessary for accurate learning. RCD and RPC recover the exact skeleton, whereas the best and worst PPC cases learn flawed skeletons (and also flawed oriented models), with high false positive and high false negative rates. This is evidence that propositionalizing relational data may lead to inaccurately learned causal models.

For oriented models, the RCD algorithm vastly exceeds the performance of all other algorithms. As the soundness result suggests, RCD achieves a compelled precision of 1.0, whereas RPC introduces orientation errors due to reasoning over the class dependency graph and missing additional separating sets. For recall, which is closely tied to the completeness result, RCD ranges from roughly 0.56 (for 1 dependency and 2 entities) to 0.94 (for 15 dependencies and 4 entities). While RPC and PPC cannot orient models with a single dependency, the relational bivariate orientation rule enables RCD to orient models using little information. RCD also discovers more of the underlying causal structure as the complexity of the domain increases, with respect to both relational structure (more entity and relationship classes) and model density.

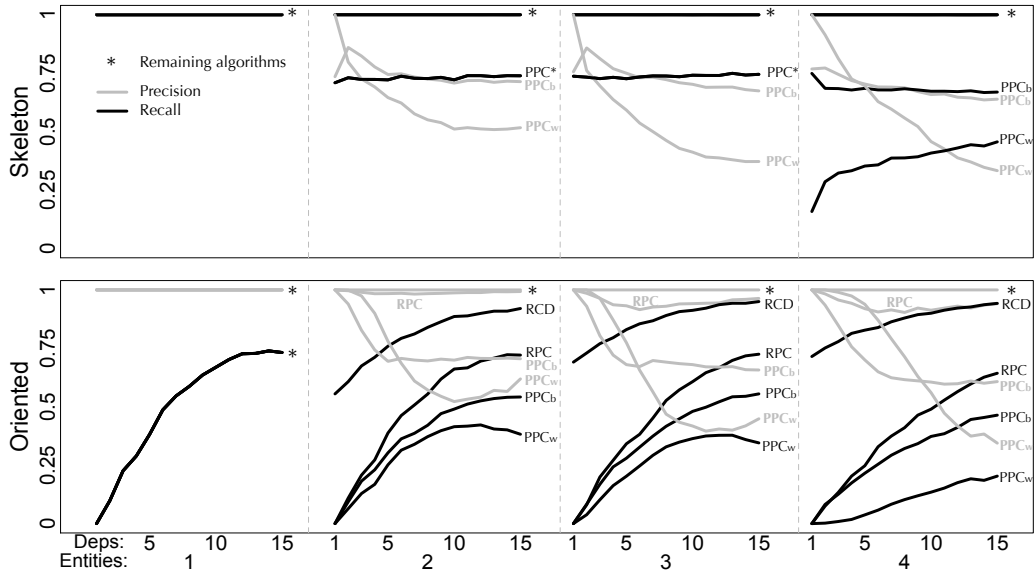

Figure 5: Skeleton and oriented precision and recall for the RCD and RPC algorithms, as well as the best and worst perspective for PPC for a baseline. Results are averaged over 1,000 models for each setting.

To quantify the unique contribution that RBO provides, we applied RBO as the *final* orientation rule in Phase II and recorded the frequency with which each edge orientation rule is activated (see Figure 6). As expected, RBO never activates for the single-entity case because all paths have cardinality ONE. For truly relational domains, RBO orients between 11% and 100% of the oriented edges. However, this does not fully capture the broad applicability of RBO. Therefore, we also recorded the frequency of each edge orientation rule when RBO is applied *first* in Phase II of RCD. In this case, for at least two entity classes, RBO orients between 58% and 100% of the oriented edges.

Finally, we recorded the number of conditional independence tests used by the RCD and RPC algorithms. RCD learns a more accurate model than RPC, but at the cost of running additional tests of independence during Phase II. Fortunately, these extra tests do not alter the asymptotic complexity of the algorithm, requiring on average 31% more tests.

### 6.2 DEMONSTRATION ON REAL DATA

We applied RCD to the MovieLens+ database, a combination of the UMN MovieLens database (www.grouplens.org); box office, director, and actor information collected from IMDb (www.imdb.com); and average critic ratings from Rotten Tomatoes (www.rottentomatoes.com). Of the 1,733 movies with this additional information, we sampled 10% of the user ratings yielding roughly 75,000 ratings. For testing conditional independence, RCD checks the signif-icance of coefficients in linear regression and uses the average aggregation function for relational variables. The RCD-generated model is displayed in Figure 7.

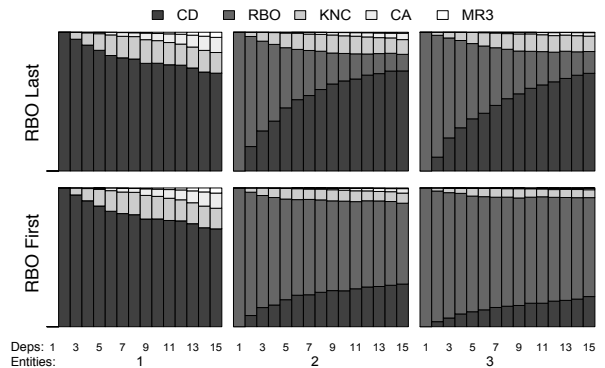

Figure 6: Frequency of edge orientation rules in RCD, with RBO last (above) and first (below).

We ran RCD with a hop threshold of 4, maximum depth of 3, and an effect size threshold of 0.01. Because constraint-based methods are known to be order-dependent (Colombo and Maathuis, 2012), we ran RCD 100 times and used a two-thirds majority vote to determine edge presence and orientation. RCD discovered 27 dependencies. One interesting dependency is that the average number of films that actors have starred in affects the number of films the director has directed—perhaps well-established actors tend to work with experienced directors. Also, note that genre is a composition of binary genre attributes.

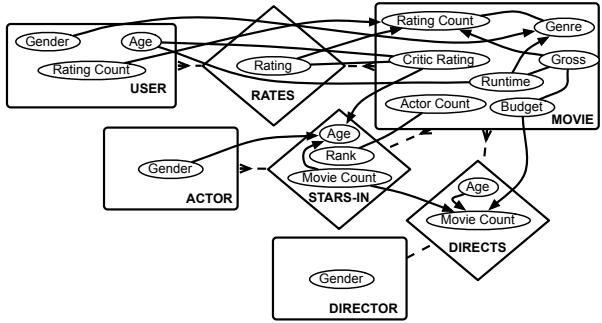

Figure 7: RCD-learned model of MovieLens+.

## 7 RELATED WORK

The ideas presented in this paper are related to three primary areas of research. First, the RCD algorithm is a constraint-based method for learning causal structure from observed conditional independencies. The vast majority of other causal discovery algorithms have focused on Bayesian networks and propositional data. The IC (Pearl, 2000) and PC (Spirtes et al., 2000) algorithms provided a foundation for all future constraint-based methods, and Meek (1995) proved these equivalent methods to be sound and complete for causally sufficient data. Additional constraint-based methods include the Grow-Shrink (Margaritis and Thrun, 1999) and TC (Pellet and Elisseeff, 2008) algorithms.

Second, RCD emphasizes learning causal *relational* models, a more expressive class of models for real-world systems. Our experimental results also indicate that propositional approaches may be inadequate to handle the additional complexity of relational data. Algorithms for learning the structure of directed relational models have been limited to methods based on search-and-score that do not identify Markov equivalence classes (Friedman et al., 1999). The RPC algorithm was the first to employ constraint-based methods to learn causal models from relational data (Maier et al., 2010), but RPC is not complete and may introduce errors due to its underlying representation. Both RPC and PRMs include capabilities to reason about relationship existence (Getoor et al., 2002); however, we currently focus on attributional dependencies and leave causes of existence as future work.

Finally, orienting bivariate dependencies, the most effective constraint used by RCD, is similar to the efforts of Shimizu et al. (2006), Hoyer et al. (2008), and Peters et al. (2011) in the propositional setting. Contrary to RBO, these techniques leverage strong assumptions on functional form and asymmetries in conditional densities to determine the direction of causality. Nonetheless, these methods could orient some of the edges that remain unoriented by RCD, given these additional distributional assumptions.

## 8 CONCLUSIONS

Relational $d$-separation and the abstract ground graph representation provide a new opportunity to develop theoretically correct algorithms for learning causal structure from relational data. We presented the relational causal discovery (RCD) algorithm and proved it sound and complete for discovering causal models from causally sufficient relational data. We introduced relational bivariate orientation (RBO), which can detect the orientation of bivariate dependencies. This leads to recall of oriented relational models over a previous state-of-the-art algorithm that is 18% to 72% greater on average. We also demonstrated RCD's effectiveness on synthetic causal relational models and demonstrated its applicability to real-world data.

There are several clear avenues for future research. RCD could be extended to reason about entity and relationship existence, and the assumptions of causal sufficiency and acyclic models could be relaxed to support reasoning about latent common causes and temporal dynamics. There are also new operators that exploit relational structure, such as relational blocking (Rattigan et al., 2011), which could be integrated with simple tests of conditional independence. Finally, RCD could be enhanced with Bayesian information, similar to the recent work by Claassen and Heskes (2012) for improving the reliability of algorithms that learn the structure of Bayesian networks.

## Acknowledgments

The authors wish to thank Cindy Loiselle for her editing expertise. This effort is supported by the Intelligence Advanced Research Project Agency (IARPA) via Department of Interior National Business Center Contract number D11PC20152, Air Force Research Lab under agreement number FA8750-09-2-0187, the National Science Foundation under grant number 0964094, and Science Applications International Corporation (SAIC) and DARPA under contract number P010089628. The U.S. Government is authorized to reproduce and distribute reprints for governmental purposes notwithstanding any copyright notation hereon. The views and conclusions contained herein are those of the authors and should not be interpreted as necessarily representing the official policies or endorsements either expressed or implied, of IARPA, DoI/NBC, AFRL, NSF, SAIC, DARPA or the U.S. Government. The Greek State Scholarships Foundation partially supported Katerina Marazopoulou.